# Title

Challenges in the application of a mortality prediction model for COVID-19 patients on an Indian cohort

# Author list

Yukti Makhija[#], Department of Biochemical engineering and Biotechnology, Indian Institute of Technology, Delhi, Hauz Khas, New Delhi-110016. bb1190067@iitd.ac.in. ORCID No: 0000-0001-8062-7206

Samarth Bhatia[#], Department of Chemical engineering, Indian Institute of Technology, Delhi, Hauz Khas, New Delhi-110016. ch1190124@iitd.ac.in. ORCID No: 0000-0002-2056-2225

Dr. Shalendra Singh, MD, DNB, DM, Associate Professor, Department of Anesthesiology and Critical Care, AFMC, Pune-411040, Maharashtra. email: drsinghafmc@gmail.com. ORCID No: 0000-0002-1112-3431

Dr. Sneha Kumar Jayaswal, MBBS. Department of Biochemical engineering and Biotechnology, Indian Institute of Technology, Delhi, Hauz Khas, New Delhi-110016. Email: drsnehakj@gmail.com

Dr. Prabhat Singh Malik, MD, DM. Associate Professor, Department of Medical Oncology, All India Institute of Medical Sciences, Ansari Nagar, New Delhi-10029. email: drprabhatsm@gmail.com

Pallavi Gupta, Department of Biological Sciences, Indian institute of Science education and research, Bhopal, Bhauri, Madhya Pradesh – 462066. Email: pgupta3005@gmail.com

Shreyas N. Samaga, MS. Department of Biochemical engineering and Biotechnology, Indian Institute of Technology, Delhi, Hauz Khas, New Delhi-110016. Email: samagashreyas@gmail.com

Shreya Johri, Department of Biochemical engineering and Biotechnology, Indian Institute of Technology, Delhi, Hauz Khas, New Delhi-110016. Email:shreyajohri16@gmail.com

Dr. Sri Krishna Venigalla, MBBS. Dept of Anesthesiology & Critical Care, Armed forces Medical College, Pune-411040, Maharashtra. Email: srikrishna9@gmail.com

Dr Rabi Narayan Hota, MD, Dept of Anesthesiology & Critical Care, Armed forces Medical College, Pune-411040, Maharashtra. Email: rabi072027@gmail.com

Dr Surinder Singh Bhatia, Mphil, DGAFMS office, Ministry of Defence, New Delhi-110010. Email: bhatia_sssss@yahoo.co.in

Dr. Ishaan Gupta*, B.E, M.Sc., Ph.D. Assistant Professor, Department of Biochemical engineering and Biotechnology, Indian Institute of Technology, Delhi, Hauz Khas, New Delhi-110016. email: ishaan@iitd.ac.in. ORCID No: 0000-0001-8934-9919

#These authors contributed equally to the article
* corresponding author (ishaan@iitd.ac.in)

Many countries are now experiencing the third wave of the COVID-19 pandemic straining the healthcare resources with an acute shortage of hospital beds and ventilators for the critically ill patients. This situation is especially worse in India with the second largest load of COVID-19 cases and a relatively resource-scarce medical infrastructure. Therefore, it becomes essential to triage the patients based on the severity of their disease and devote resources towards critically ill patients. Yan et al. [1] have published a very pertinent research that uses Machine learning (ML) methods to predict the outcome of COVID-19 patients based on their clinical parameters at the day of admission. They used the XGBoost algorithm, a type of ensemble model, to build the mortality prediction model. The final classifier is built through the sequential addition of multiple weak classifiers. The clinically operable decision rule was obtained from a 'single-tree XGBoost' and used lactic dehydrogenase (LDH), lymphocyte and high-sensitivity C-reactive protein (hs-CRP) values. This decision tree achieved a 100% survival prediction and 81% mortality prediction. However, these models have several technical challenges and do not provide an out of the box solution that can be deployed for other populations as has been reported in the "Matters Arising" section of Yan et al. Here, we show the limitations of this model by deploying it on one of the largest datasets of COVID-19 patients containing detailed clinical parameters collected from India.

Our dataset was collected as a part of a retrospective study that was conducted at two centers for COVID-19 in New Delhi, namely, Sardar Vallabhbhai Patel Covid Hospital and PM Cares Covid Care Hospital. Sardar Vallabhbhai Patel Covid hospital took the lead for this study and coordinated with the other centre for data collection and analysis. The study period was from 13th July 2020 to 14th October 2020. Every case with a confirmed infection who was admitted at any of the participating centres between this period, for treatment of COVID-19 infection was diagnosed by a Rapid Antigen Test or RT-PCR testing of a nasal/throat swab sample. Patients of all age groups were included in the study. Most of the data was retrieved from retrospectively maintained medical records. The clinical classification for COVID-19 severity was defined according to the Ministry of Health and Family Welfare definition (MoHFW, Government of India). All the hospitalized cases were followed up till discharge or death during COVID-19 illness. The primary criteria for discharge in mild/moderate hospitalized cases was the resolution of symptoms and a minimum stay of 10 days in the absence of follow-up RT-PCR negativity. Patients with severe infection were discharged only after clinical recovery and with a negative RT-PCR on repeat swab after resolution of symptoms.

We tested how the model published by Yan et al performs on the above dataset of Indian patients. Although we collected data from 841 patients, most of these patients did not undergo all the clinical tests and, hence we did not have the clinical parameters used to sort patients based on the decision rule presented in the paper. Many of these parameters were not collected and this presented one of the biggest challenges in deploying ML-models in a resource-scarce environment such as India, where these models can be most useful in terms of managing patient load given poor doctor: patient ratio of 1:1456[2]. Only 120 patients (Table 1) had all the parameters measured necessary in the Yan et al study, out of which 95 (79.17 %) were discharged, and 25 (20.83%) were deceased. Of the 95 patients who had recovered, 23 (19.17%) patients experienced mild symptoms, 42 (35%) moderate, and 30 (25%) severe. The overall survival prediction accuracy was 65.26%, and the mortality prediction accuracy was 88%. Individual survival prediction in the case of mild, moderate, and severe patients was 91.3%, 73.81%, and 33.33%, respectively (Figure 1).

Besides the fewer number of parameters measured from each patient, we uncovered a major challenge. As reported previously by Reeve et al.[3], there are two kits used in practice to measure LDH levels based on conversion of lactate to pyruvate or vice versa. We used the later kit which has a reference range of values of 240-480 $Ul^{-1}$. While Yan et al most likely used the former kit with a reference range of values of 135-250 $Ul^{-1}$. Hence, we had to normalize the LDH values in our dataset as shown previously[4]. This again reiterates the previous concerns [3] about deploying machine learning models that may require the knowledge of reference range values of the biochemical tests performed and appropriate normalization between datasets.

We conclude that on Indian patients the decision rule by Yan et al is a good predictor of mortality but underperforms in predicting survival of COVID-19 patients. Further, it particularly underperforms in predicting severity of infection which is the key prediction necessary for effective patient triage and resource allocation. Our results present an opposite trend from a similar replication study by Quanjel et on a Dutch cohort[5], who report good performance of the decision rule to predict patient severity but not mortality. The discrepancy in replicating the results from Yan et al could possibly be explained by the differences in access to medical infrastructure between the population used for training the models, by demography (age, gender ratio distribution), or by the differences in reference ranges of the selected parameters due to population genetics. An observation further strengthened the conjecture regarding biases due to population genetics. The minor allele frequencies of rs7305678, an allele that is significantly associated with LDH serum levels[6], were dramatically different in European, East Asian, and South Asian (Indian) populations at 0.48, 0.13, 0.21 respectively according to the 1000 Genomes Project[7,8]. We propose that other yet genetic factors may also influence quantitative traits such as biochemical parameter values between different populations. These differences may also manifest as differences in patient mortality rates which are quite different between countries such as India has a mortality rate at about 1.5%, while China has a mortality rate at about 3.8% [9].

Therefore, our results suggest that machine learning models being developed to predict patient outcomes, especially those predicting large scale clinical manifestation in pandemics, need to take into account the biases in the collected feature values such as technical variability in biochemical parameters, population genetics, demography, and other socio-economic factors in healthcare.

**Specific Authors Contribution:**
IG, SB, YM, SJ designed the study.
SKJ, PG, SNS, RNH, SSB, PSM, SKV and SS collected and curated the data.
SB, YM performed data analysis.
IG, SS, SB, YM contributed in writing the Manuscript.
All authors read and approved the final manuscript.

**Confidentiality statement and ethics**

The study protocol was reviewed and approved by institute ethics committees at both the participating centres through Institute ethics committee File No. IEC/320/325 to SS. Consent waiver was granted given the retrospective nature of analysis and emergent nature of the pandemic. Confidentiality was maintained by the de-identification of data. All analysis was performed on de-identified data.

**Acknowledgment**

We would like to acknowledge the nursing staff and medical professionals who have tirelessly worked to alleviate the suffering caused by this Pandemic and facilitated the collection of good quality data for this publication.

**Financial support & sponsorship**

This research did not receive any specific grant from funding agencies in the public, commercial, or not-for-profit sectors.

**Conflicts of Interest**

The authors have no conflict of interest.

**Figures and Figure Legends**

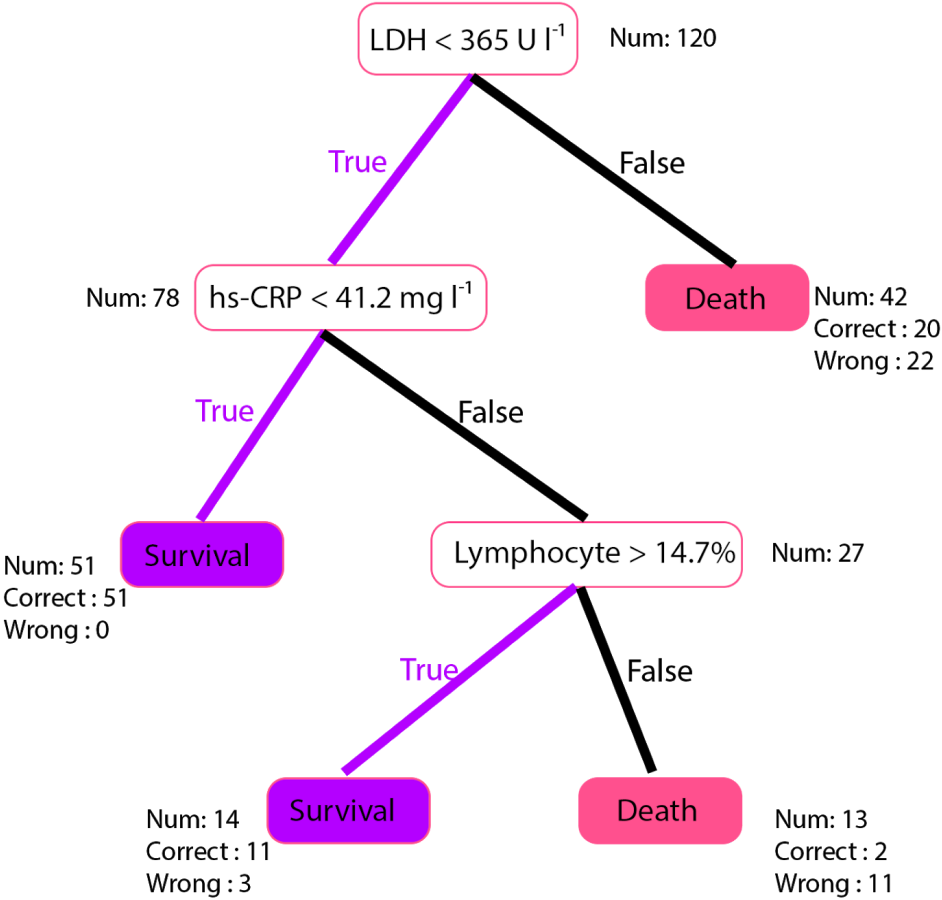

**Figure 1:** Evaluation of the Yan et al.'s decision rule tree populated with our dataset

**Tables**

| Features/Characteristics | Value |
|---|---|
| Age \| Average (Std Dev), Median | 56.66 (15.18), 57 |
| Gender \| Number, %age of total | |
|     Male | 85 (70.83%) |
|     Female | 35 (29.16%) |
| Outcome \| Number (%age of total) | |
|     Survival Rate | 95 (79.17%) |
|     Mortality Rate | 25 (20.83%) |
| Clinical Data \| Average, Median | |
|     Lactate Dehydrogenase (LDH) | 364.76 U $L^{-1}$, 287.74 U $L^{-1}$ |
|     High sensitivity C-Reactive Protein | 42.30 mg $L^{-1}$, 34.39 mg $L^{-1}$ |
|     Lymphocytes (%) | 18.45%, 16% |

**Table 1**: Description of our dataset showing statistical measures of central tendency